\documentclass{Interspeech}
\newcommand{\bolditalic}[1]{\textbf{\textit{#1}}}

\interspeechcameraready 

\title{Zero-Shot Streaming Text to Speech Synthesis with Transducer and Auto-Regressive Modeling}

\author[affiliation={1,2}]{Haiyang}{Sun}
\author[affiliation={1}]{Shujie}{Hu}
\author[affiliation={1}]{Shujie}{Liu}
\author[affiliation={1}]{Lingwei}{Meng}
\author[affiliation={1}]{Hui}{Wang}
\author[affiliation={2}]{Bing}{Han}
\author[affiliation={1}]{Yifan}{Yang}
\author[affiliation={1}]{Yanqing}{Liu}
\author[affiliation={1}]{Sheng}{Zhao}
\author[affiliation={1}]{Yan}{Lu}
\author[affiliation={2}]{Yanmin}{Qian}

\affiliation{}{Microsoft Corporation}{}
\affiliation{}{Shanghai Jiao Tong University}{}
\email{sunhaiyang@sjtu.edu.cn}

\usepackage{comment}
\usepackage{booktabs}
\usepackage[table]{xcolor}

\begin{document}

\maketitle

\begin{abstract}
    Zero-shot streaming text-to-speech is an important research topic in human-computer interaction. Existing methods primarily use a lookahead mechanism, relying on future text to achieve natural streaming speech synthesis, which introduces high processing latency. To address this issue, we propose SMLLE, a streaming framework for generating high-quality speech frame-by-frame. SMLLE employs a Transducer to convert text into semantic tokens in real time while simultaneously obtaining duration alignment information. The combined outputs are then fed into a fully autoregressive (AR) streaming model to reconstruct mel-spectrograms. To further stabilize the generation process, we design a Delete \(\langle Bos \rangle\) Mechanism that allows the AR model to access future text introducing as minimal delay as possible. Experimental results suggest that the SMLLE outperforms current streaming TTS methods and achieves comparable performance over sentence-level TTS systems. Samples are available on \url{shy-98.github.io/SMLLE_demo_page/}.
    
\end{abstract}

\section{Introduction}

In recent years, large language models (LLMs) have made remarkable advancements across various domains, including natural language processing (NLP) \cite{achiam2023gpt, dubey2024llama} and computer vision (CV) \cite{ramesh2021zero, radford2021learning}. Similarly, the field of speech synthesis has seen significant progress in autoregressive (AR) language modeling, with both discrete codec-based \cite{wang2023neural, zhang2023speak, chen2024vall} and continuous mel-spectrogram-based approaches \cite{meng2024autoregressive}. By harnessing the in-context learning capabilities and scalability of AR models, these methods have achieved exceptional performance in zero-shot text-to-speech (TTS) synthesis.
However, to synthesize high-quality speech, LLM-based zero-shot TTS models typically process entire sentences as input before generating speech frame by frame. This approach results in significant latency and limits the ability to handle very long texts efficiently, hindering the model's ability of real-time generation, and negatively impacting the user interaction experience \cite{defossez2024moshi,xie2024mini,fang2024llama,tang2023salmonn,chu2023qwen,hu2024wavllm}. 

To address this issue, zero-shot streaming TTS models aim to enable real-time speech generation while preserving in-context learning capabilities.
Existing streaming TTS methods can be broadly categorized into two approaches. The first is chunk-level generation \cite{dang2024livespeech}, where long texts are divided into smaller segments, and speech is synthesized separately for each chunk. While this lookahead mechanism enhances synthesis quality, it inevitably introduces latency proportional to the chunk size.
The second approach is frame-by-frame generation \cite{chen2021speech, kim2023transduce, lee2024high, bataev2025tts}, which leverages the real-time generation capabilities of the Transducer model to minimize latency. For instance, Speech-T \cite{chen2021speech} employs a Transducer to directly model text-to-speech alignment, improving performance by constraining the alignment path during training. However, while it performs well in single-speaker scenarios, it lacks zero-shot capability for generating speech from unseen speakers.
Other approaches \cite{kim2023transduce, lee2024high} use a Transducer to generate semantic tokens, text-related representations that are decoupled from speech. These tokens are then converted into speech using separate models. Specifically, \cite{kim2023transduce} employs a VITS-based non-autoregressive model for speech reconstruction, while \cite{lee2024high} adopts a cross-attention based seq2seq model(Grouped Masked Language Model). Although both methods exhibit strong zero-shot capability, their final speech generation models operate at the sentence level rather than in a streaming manner, negating the low-latency benefits that the Transducer's streaming semantic generation could provide.

\begin{figure}[t]
    \centering
    \includegraphics[width=0.48\textwidth]{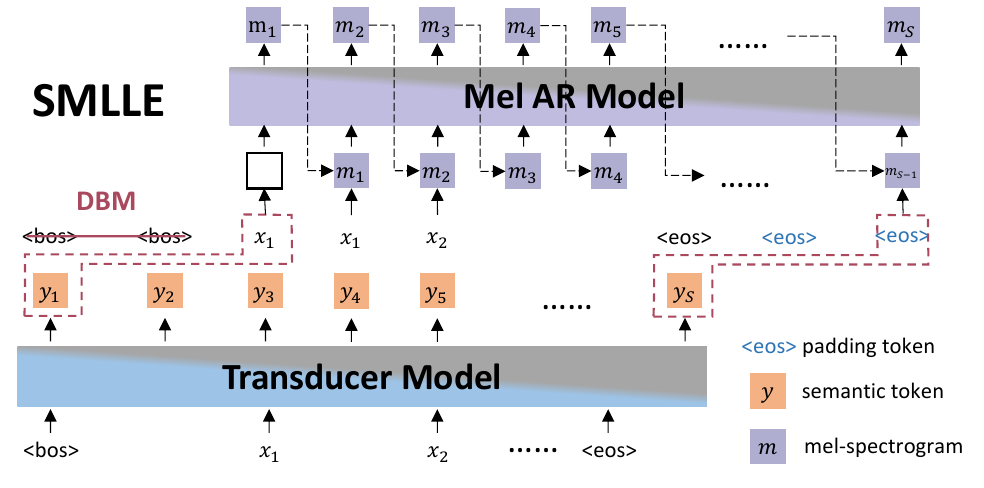}
    \vspace{-7mm}
    \caption{The overview of SMLLE.}
    \vspace{-7mm}
    \label{fig:overview}
\end{figure}

To build a zero-shot streaming TTS model, in this paper, we propose a new framework, SMLLE, which generates mel-spectrograms streamingly with very little latency.  As shown in Figure \ref{fig:overview}, SMLLE has two key components: (1) A Transducer based streaming model to generate the semantic tokens, based on the text.
(2) An AR streaming model to generate the mel spectrum sequence for the final speech reconstruction, based on the semantic token sequence generated by the first component and the duration-aligned text with the duration information from the Transducer model.
Our contributions can be summarized:
\begin{enumerate}
\item We propose the \textbf{first work of zero-shot streaming TTS model in frame-by-frame mode, SMLLE}. Specifically, it uses a Transducer to convert the text into a sequence of semantic token in real time. At the same time, a fully AR model converts these semantic tokens and texts into mel-spectrograms, frame by frame.
\item We introduce \textbf{a novel ``Delete \(\langle Bos \rangle\) Mechanism'' (DBM)} for SMLLE. This mechanism allows the model to access necesssary future text introducing as minimal delay as possible, thereby improving the quality of speech reconstruction.
\item The experimental results show that SMLLE achieves \textbf{ performance on par with zero-shot non-streaming TTS models}. 
Specifically, in terms of objective metrics, a WER of 6.4\% and a speaker similarity score of 0.54 are obtained using SMLLE, which are comparable to that of VALL-E \cite{wang2023neural}. 

\end{enumerate}

\section{SMLLE}

SMLLE generates high-quality speech using a two-stage modeling. In the Transducer Stage, SMLLE converts the text sequence into a semantic tokens sequence in a streaming manner. In the Autoregressive Stage, it uses the semantic tokens and texts to reconstruct mel-spectrograms frame by frame.

\subsection{Transducer Stage}

The Transducer is proposed to model the monotonic alignment transformation between the input sequence and the output sequence. It consists of three components: 1) an encoder for the input sequence encoding, 2) a predictor for the output sequence modeling, and 3) a Joint-Net that predicts the next output by combining the states of encoder and predictor. Following previous work, the Transducer model aligns text with semantic tokens, which are text-related representations that are decoupled from speech. Specifically, sementic tokens are the first layer codecs extracted from the SpeechTokenizer \cite{zhang2023speechtokenizer}.


We denote the text sequence \( \bolditalic{X} = \{ \langle bos \rangle, \bolditalic{x}_t, \langle eos \rangle \}^T_{t=1} \) as the input, \( \langle bos \rangle \) and \( \langle eos \rangle \) as \( \bolditalic{x}_0 \) and \( \bolditalic{x}_{T+1} \) respectively, the semantic tokens \( \bolditalic{Y} = \{ \bolditalic{b} , \bolditalic{y}_s\}^S_{s=1}\) as the target. \( \bolditalic{b} \) is a shared initialization state, which we denote as \( \bolditalic{y}_0 \) in the following. The Transducer model gradually considers whether it can predict the \(\bolditalic{y}_{j+1}\) given the \(\bolditalic{x}_{0:i}\) and the \(\bolditalic{y}_{0:j}\), where \(0 \le i \le T+1, 0 \le j<S\); Transducer will output a special token  \( \langle blank \rangle \) if it cannot make a confident prediction, and will consider the \(\bolditalic{x}_{0:i+1}\) to gather future text information; When \(i=T+1\), the model outputs \( \langle blank \rangle \), indicating the end of the generation.

\begin{figure}[t]
    \centering
    \vspace{-3mm}
    \includegraphics[width=0.25\textwidth]{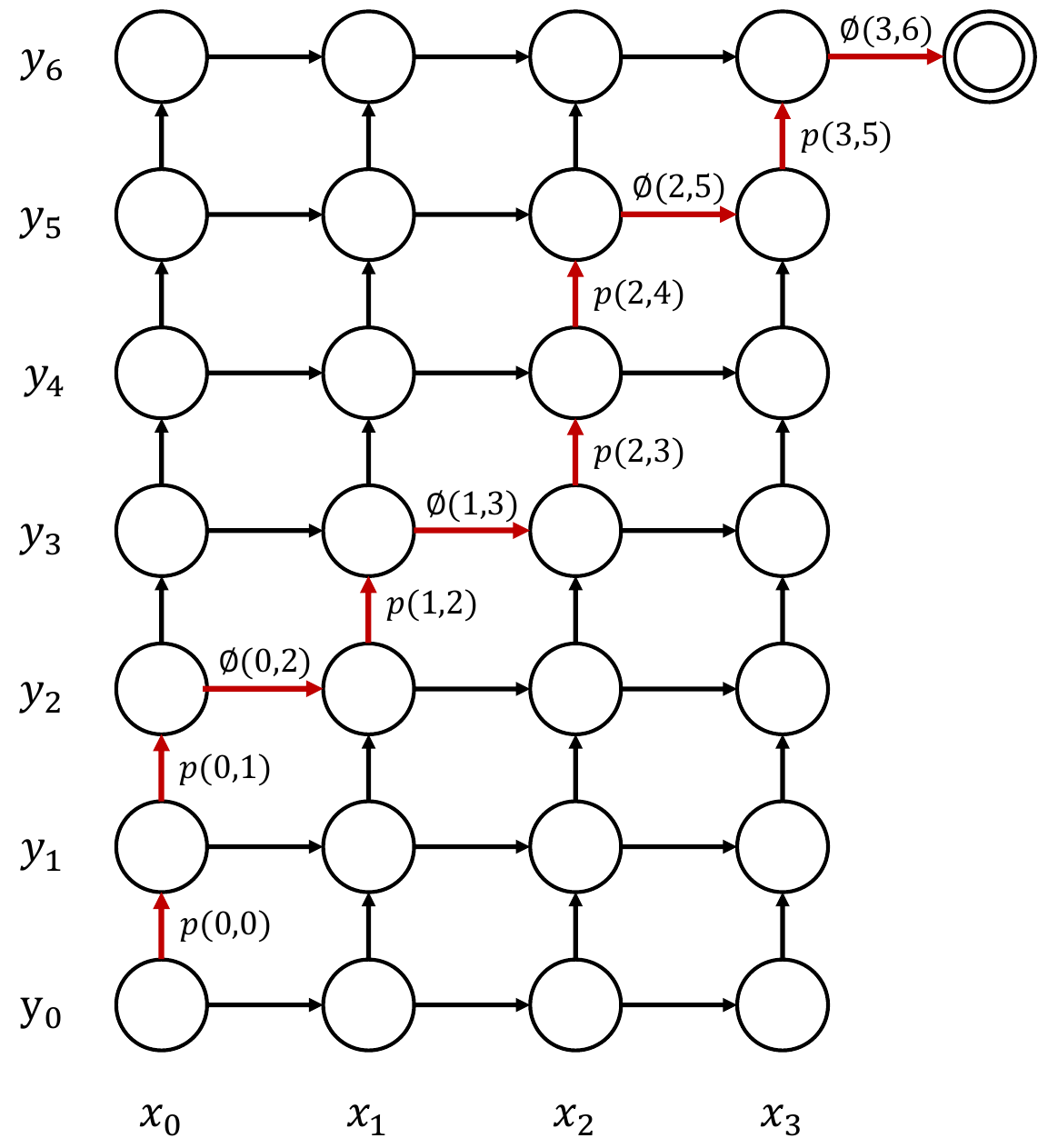}
    \vspace{-3mm}
    \caption{The probabilistic path graph of the Transducer, where the red paths represent the possible alignment paths.}
    \vspace{-5mm}
    \label{fig:transducer_grid}
\end{figure}

Figure \ref{fig:transducer_grid} presents a grid that shows all possible alignment paths during the Transducer modeling process. At each node, the Transducer calculates the emission probability \( P(\bolditalic{y}_{j+1}|\bolditalic{x}_{0:i},\bolditalic{y}_{0:j}) \), represented by the vertical arrows \( p(i,j) \), and the wait probability \( P(\langle blank  \rangle|\bolditalic{x}_{0:i},\bolditalic{y}_{0:j})) \), represented by the horizontal arrows \( \varnothing(i,j) \). During training, it marginalizes over all legal alignments \( \mathcal{\bolditalic{A}} \) between \( \bolditalic{X} \) and \( \bolditalic{Y} \), and minimizes the negative log probability of the conditional distribution:

\begin{equation}
    \mathcal{L}_T=- \log{P(\bolditalic{Y}|\bolditalic{X})}= - \log{\sum_{\alpha \in \mathcal{F}^{-1}(\bolditalic{Y}) }{P(\alpha|\bolditalic{X})}}.
\end{equation}
Here, \(\alpha\) represents all possible monotonic alignment paths. \(F^{-1}\) is the inverse function of \(F\), and the function \(F^{-1}(\bolditalic{Y})\) generates all possible paths by inserting \( \langle blank \rangle \) into \(\bolditalic{Y}\). The sum of the probabilities of all paths is calculated using an efficient forward algorithm. The probability of reaching the node \(\alpha(i,j)\) can be expressed as:
\begin{equation}
    \alpha(i, j) = \alpha(i - 1, j) \cdot p(i - 1, j) + \alpha(i, j - 1) \cdot p(i, j - 1),
\end{equation}
where the initial condition is \(\alpha(0, 0) = 1\). During the inference phase, the model first receives the information \( \bolditalic{x}_{0:0} \) and \( \bolditalic{y}_{0:0} \). It then streams to generate and receive subsequent semantic tokens and texts. Meanwhile, \( \bolditalic{X} \) can be replicated based on the vertical path to obtain the duration-aligned text \( \bolditalic{X}' \), which will be used as the input of the DBM.

\subsection{Autoregressive Stage}

At this stage, SMLLE models the mel-spectrograms autoregressively using semantic tokens, together with the duration-aligned text to provide essential information.
This stage not only leverages the in-context learning capabilities of autoregressive modeling to achieve strong zero-shot performance, but also avoids the high-latency issues by using a streaming generation approach. Additionally, using the Delete ⟨Bos⟩ Mechanism strategy, the AR model achieves better performance while minimizing unnecessary latency as much as possible.

We denote the mel-spectrograms as \( \bolditalic{M} = \{ \bolditalic{m}_s \}_{s=1}^{S} \), the duration-aligned text as \( \bolditalic{X}' = \{ \bolditalic{x}'_s \}_{s=1}^{S} \), and the semantic tokens without  initialization state \( \bolditalic{Y} = \{ \bolditalic{y}_s \}_{s=1}^{S} \). The sequence lengths of \( \bolditalic{M} \), \( \bolditalic{X}' \) and \( \bolditalic{Y} \) are consistent. 

\begin{figure}[t]
    \centering
    \vspace{-3mm}
    \includegraphics[width=0.48\textwidth]{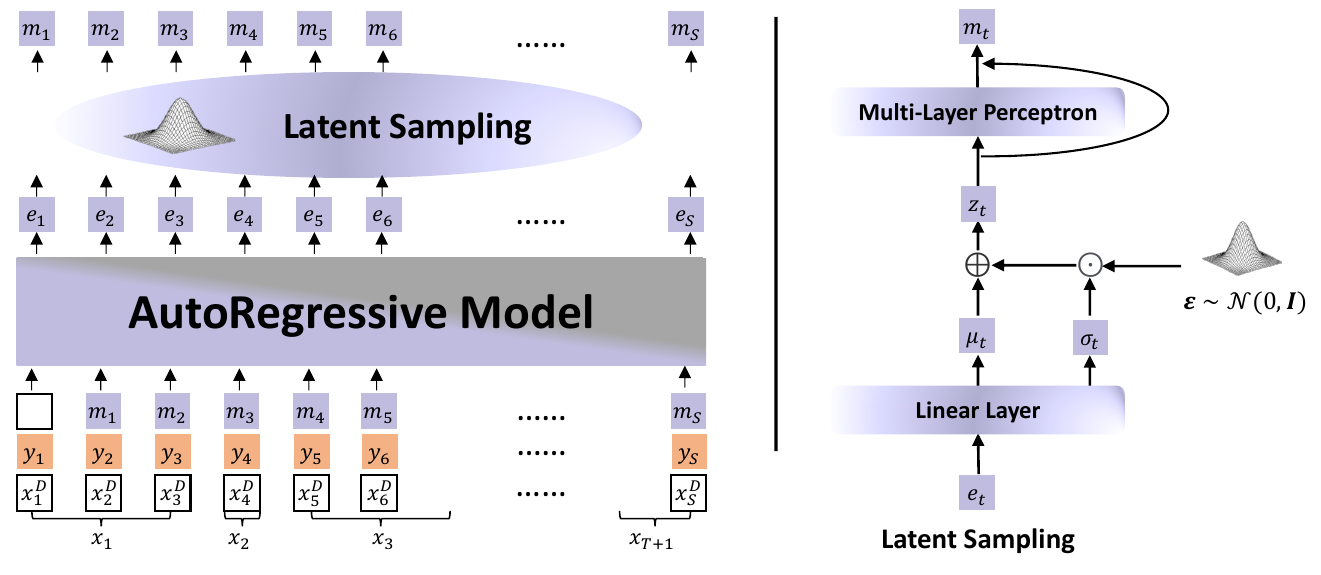}
    \vspace{-7mm}
    \caption{The AR model utilizes both duration-aligned text and semantic tokens to provide essential information for mel-spectrogram reconstruction. The latent sampling module is introduced to enhance the model's generalization capability. The relationship between \(x^D\) and \(x\) is indicated in the brackets.}
    \vspace{-5mm}
    \label{fig:ARModel}
\end{figure}

\subsubsection{Delete \(\langle Bos \rangle\) Mechanism}

In each speech segment, there is a brief period of silence at the beginning and end to make the interaction more natural. In the Transducer Stage, we represent these silent intervals with the special tokens \( \langle bos \rangle \) and \( \langle eos \rangle \). 

However, the \( \langle bos \rangle \) token does not have a clear phonetic meaning. Therefore, the \( \langle bos \rangle \) text token is unnecessary for AR stage. By removing \( \langle bos \rangle \), the model can access future text earlier without introducing much latency, thus making the streaming generation process more stable and improving performance. Of course, an equal number of \( \langle eos \rangle \) need to be appended to the end of the sequence to ensure consistent sequence length. After applying the DBM, we obtain the final input, denoted as \(\bolditalic{X}^D\). The specific calculation process can be described as follows:

\begin{equation}
    \mathbf{X}^D = \text{remove}_{bos}({\mathbf{X}'}) \parallel \underbrace{\langle \text{eos} \rangle \dots \langle \text{eos} \rangle}_{\# \langle \text{bos} \rangle} .
\end{equation}

\subsubsection{Architecture}

Following MELLE~\cite{meng2024autoregressive}, we use the mel-spectrogram as both the input and output of the model. Building on this, we further set the text and semantic tokens as inputs to enable better generation performance. At the \( t \)-th step, the model receives the \( t \)-th semantic token \( \bolditalic{y}_t\), text token \( \bolditalic{x}^D_t\) and the \( t \)-1-th frame of the mel-spectrogram \( \bolditalic{m}_{t-1}\), to predict the \( \bolditalic{m}_{t}\), by maximizing the following distribution:

\begin{equation}
    p(\bolditalic{m} \mid \bolditalic{x}^D; \bolditalic{y}; \theta) = \prod_{t=1}^{T} p(\bolditalic{m}_t \mid \bolditalic{m}_{<t}; \bolditalic{x}^D_{\leq t}; \bolditalic{y}_{\leq t}; \theta),
\end{equation}
where $\theta$ represents the parameters of AR model.

To enhance speech diversity and the model's generalization ability, we adopt the latent sampling module in MELLE. As shown in Figure \ref{fig:ARModel}, the output hidden state of the AR model at step \( t \) is denoted as \( \bolditalic{e}_{t}\). This hidden state  \( \bolditalic{e}_{t}\) is sent to a linear layer to predict the mean vector \( \bm{\mu}_{t} \) and the log variance \( \log\bm{\sigma}_t^2 \). Then, independent reparameterization sampling for each dimension yields \( \bolditalic{z}_t \), following a multivariate Gaussian distribution:
\begin{align}
\bm{z}_t = \bm{\mu}_t + \bm{\sigma}_t \odot \bm{\epsilon}, \quad \text{where} \quad \bm{\epsilon} \sim \mathcal{N}(0, \bm{I}), 
\end{align}
and the probability density function can be defined
\begin{align}
\label{eq:gaussian}
&p_\theta(\bm{z}_t \mid \bm{e}_t) = \mathcal{N}(\bm{z}_t \mid \bm{\mu}_t, \mathrm{diag}(\bm{\sigma}_t^2)).
\end{align}

The latent sampling module uses the reparameterization technique, making it differentiable. The latent variable $\bm{z}_t$ is then processed through a multi-layer perceptron (MLP) that includes residual connections, generating the corresponding mel-spectrogram $\bm{m}_{t}$. Since the $\bm{m}_{t}$ cannot observe future information during streaming generation, we no longer use Post-Net to refine all mel-spectrograms. Instead, we directly use the results generated by the MLP as the final output.

In terms of training objectives, we follow MELLE \cite{meng2024autoregressive} and use Regression Loss $\mathcal{L}_{\text{reg}}$, KL Divergence Loss $\mathcal{L}_{\text{KL}}$, and Spectrogram Flux Loss $\mathcal{L}_{\text{flux}}$ as the loss functions. Since the duration of the generated speech is determined by the Transducer model, we do not use Stop Prediction Loss.
The final training loss of the AR model is the weighted sum of these three types of losses:
\begin{align}
\mathcal{L}_{\text{AR}} = \mathcal{L}_{\text{reg}} + \lambda \mathcal{L}_{\text{KL}} + \beta\mathcal{L}_{\text{flux}},
\end{align}
where $\lambda$ and $\beta$ are weights.

\begin{table}[h]
\caption{Performance of different systems on the \textbf{LibriSpeech} test-clean dataset. All systems, except SMLLE, are sentence-level TTS systems. The results are taken from \cite{meng2024autoregressive}. \textbf{R5} indicates the result obtained by repeating the Transducer sampling five times and selecting the best performance.}
\label{tab:librispeech}
\centering
\vspace{-3mm}
\begin{tabular}{lccc} 
\toprule
\textbf{System}                                      & \textbf{WER-C} & \textbf{WER-H} & \textbf{SIM}  \\ 
\midrule
GroundTruth                                          & 1.61           & 2.15           & -             \\
GroundTruth(vocoder)                                 & 1.58           & 2.23           & 0.734         \\ 
\midrule
\multicolumn{4}{c}{\cellcolor{gray!25}Sentence-level TTS systems}                                                         \\
ELLA-V \cite{song2024ellav}         & 7.15           & 8.90           & 0.331         \\
VALLE-R \cite{han2024valler}        & 3.18           & 3.97           & 0.395         \\
RALL-E \cite{xin2024ralle}          & 2.5            & 2.8            & -             \\
CLaM-TTS \cite{kim2024clamtts}      & -              & 5.11           & 0.538         \\
VALL-E \cite{wang2023neural}        & -              & 5.9            & 0.580         \\
VALL-E 2 \cite{chen2024valle2}      & 1.5            & 2.44           & 0.678         \\
MELLE \cite{meng2024autoregressive} & 1.47           & 2.10           & 0.664         \\
Voicebox \cite{le2024voicebox}      & -              & 1.9            & 0.681         \\ 
\midrule
\multicolumn{4}{c}{\cellcolor{gray!25}Frame-by-frame Streaming TTS systems}                                               \\
SMLLE                                                & 5.14           & 6.37           & 0.516         \\
SMLLE-R5                                             & 1.03           & 1.67           & 0.578         \\
\bottomrule
\end{tabular}
\vspace{-3mm}
\end{table}

\section{Experimental Setup}

\subsection{Training Datasets}

We train SMLLE on the LibriSpeech dataset \cite{panayotov2015librispeech}. It contains approximately 960 hours of English speech. For text, we use eSpeak 
for phoneme extraction. For the AR model, we extract 80-dimensional log-magnitude mel-spectrograms as the target.

\subsection{Experimental Settings}
\textbf{Model Configurations} For the Transducer model, we follow the setup of the Wenet\footnote{https://github.com/wenet-e2e/wenet}.
We use Conformer as the text encoder and a 2-layer LSTM as the speech encoder. Additionally, we employ two linear layers as the JointNet. 
For the AR model, similar to MELLE~\cite{meng2024autoregressive}, the mel-spectrogram is encoded through three linear layers at the input, with a dropout rate of 0.5. 
The Transducer's text encoder and AR model both contain 12 blocks. The hidden embedding dimension is 1024. Each block has 16 attention heads, a feed-forward network dimension of 4,096, and a dropout rate of 0.1.



For the vocoder, 
we train a checkpoint from scratch using the open-source BigVGAN-V2 
on LibriSpeech. To accommodate the SpeechTokenizer, we resample the input to a 16 kHz sampling rate. The upsample rates are set to [5, 4, 2, 2, 2, 2], and the upsampling kernel sizes are set to [9, 8, 4, 4, 4, 4]. The hop size for mel-spectrogram is set to 320. The model is updated for a total of 400k steps with a segment size of 81920.

\textbf{Training Details} 
For the Transducer model, 
we use the Adam optimizer with a learning rate of 0.001 and a warm-up period of 25,000 steps. The model achieves the lowest RNN-T loss after training for 12 epochs. 
For the AR model, 
we use the Adam optimizer with a learning rate of 0.0005 and a warm-up period of 30,000 steps. The \(\beta\) is set to 0.5. The \(\lambda\) is set to 5e-2 and only begins to affect model training after 10,000 steps.

\subsection{Evaluation Settings}

We use the test-clean dataset from LibriSpeech \cite{panayotov2015librispeech} and LibriTTS \cite{zen2019libritts} for evaluation, ensuring that the speakers in the datasets are not included in the training data. Based on recent studies, we select speech segments from LibriSpeech that are between 4 and 10 seconds long, and from LibriTTS, we choose segments ranging from 3 to 10 seconds. We test the model in a cross-sentence inference mode, using reference speech and its transcription from the same speaker as prompts.
\textbf{It is important to note that} the inference in the Transducer Stage does not require a prompt, utilizing a top-k sampling strategy with k=15. During the AR stage inference, dropout is applied to the input mel-spectrograms, and the prompt does not require a DBM.




We use the Conformer-Transducer \textbf{(WER-C)} and HuBERT-Large \textbf{(WER-H)} ASR models for speech recognition and to calculate the Word Error Rate (WER). We calculate speaker similarity \textbf{(SIM)} using speaker embeddings extracted by the WavLM-TDNN model. Through a crowdsourcing platform, 12 native speakers evaluate the speech quality \textbf{(MOS)} and 6 native speakers evaluate the similarity to the ground truth \textbf{(CMOS)} for the generated samples. The First Token Latency \textbf{(FTL)} is also calculated to compare the streaming generation efficiency of different models, and the \( d_{text} \) represents the number of text inputs that need to wait, and \( d_{model} \) represents the number of computational steps taken by the model.

\section{Results and Discussion}


\subsection{Main Results}

\par
\noindent
\textbf{Objective Evaluation.}
As shown in Table \ref{tab:librispeech}, the proposed frame-by-frame streaming TTS model, SMLLE achieves performance comparable to existing sentence-level TTS systems, while operating in real time.
Specifically, in terms of speaker similarity (SIM), SMLLE outperforms ELLA-V and VALLE-R, with improvements of 0.185 and 0.121, respectively, and achieves comparable results to CLaM-TTS and VALL-E, with only marginal differences of -0.022 and -0.064, respectively. 
Regarding WER, The WER-H of SMLLE is still better than that of ELLA-V with reductions of 2.53\% and only increases 
by 0.47\% compared to VALL-E. Such performance indicates the superior zero-shot capability and robustness of the SMLLE.

To further explore SMLLE's potential, five samplings on the outputs of the Transducer are conducted to ensure the best performing of semantic token prediction and alignment. As the results of ``SMLLE-R5'' shown, significant performance improvements are obtained. Specifically, the WER performance of SMLLE surpasses the current state-of-the-art systems, MELLE and Voicebox. Additionally, its SIM performance improves by 0.062, approaching the level of VALL-E.

A comparative analysis of SMLLE against existing streaming TTS systems was conducted, with results in Table \ref{tab:libritts}. LiveSpeech2-chunk \cite{dang2024zero} requires observing 4 to 8 future text chunks. Although it achieves better performance, its dependency on future text inevitably introduces latency. The first speech frame can only be generated after waiting for $4d_{text} + d_{model}$ to $8d_{text} + d_{model}$.
When restricted to frame-by-frame processing mode like SMLLE, LiveSpeech2-limit's performance degrades significantly. In contrast, SMLLE achieves excellent results (6.66\% vs. 40.7\%), while FTL also reaches $d_{text} + d_{model}$. The similar improvement after five-sampling with the Transducer further confirms the potential of SMLLE.

\par
\noindent
\textbf{Subjective Evaluation.} We sample one speech from each speaker in the LibriSpeech test-clean set, resulting in 40 test cases to conduct subjective evaluations. The results in Table \ref{tab:librispeech_mos} demonstrate that the proposed streaming TTS model, SMLLE outperforms both sentence-level YourTTS and VALL-E models in terms of MOS, and achieves superior performance on CMOS metric compared to both sentence-level YourTTS and MELLE models. These results confirm SMLLE's ability to generate highly natural and high-quality synthesized speech.

\begin{table}[t]
\caption{Performance of different streaming TTS systems on the \textbf{LibriTTS} test-clean dataset. The results are taken from \cite{dang2024zero}. \textbf{chunk} indicates that the model can observe 4 to 8 future texts, while \textbf{limit} means the model can only observe the next text. SMLLE reports the WER-C results.}
\label{tab:libritts}
\centering
\vspace{-3mm}
\resizebox{\linewidth}{!}{
\begin{tabular}{l@{\hspace{0.2cm}}c@{\hspace{0.15cm}}c@{\hspace{0.15cm}}c@{\hspace{0.15cm}}}
\toprule
\textbf{System}               & \textbf{WER} & \textbf{SIM} & \textbf{FTL} \\ \midrule
LiveSpeech2-chunk \cite{dang2024zero}   & 3.1   & 0.617 & $(4 \sim 8) d_{text}+d_{model}$ \\
LibeSpeech2-limit \cite{dang2024zero}   & 40.7  & 0.582 & $ d_{text}+d_{model}$ \\
 \midrule
SMLLE                & 6.66  & 0.528 & $ d_{text}+d_{model}$ \\
SMLLE-R5             & 1.22 & 0.603 & - \\ \bottomrule
\end{tabular}
}
\vspace{-3mm}
\end{table}

\begin{table}[t]
\caption{Subjective evaluation results for 40 speakers from the LibriSpeech test-clean dataset. The \textbf{boldface} indicates the best result, and the \underline{underline} denotes the second best.}
\label{tab:librispeech_mos}
\centering
\vspace{-3mm}

\begin{tabular}{l|c@{\hspace{0.2cm}}|c@{\hspace{0.2cm}}c@{\hspace{0.2cm}}c@{\hspace{0.2cm}}|c@{\hspace{0.2cm}}}
\toprule
\textbf{System}       & GT  & YourTTS & VALL-E & MELLE & SMLLE  \\ 
\midrule
\textbf{MOS} &3.61 & 2.15 & 3.02 & \textbf{3.39} & \underline{3.18}\\
\midrule
\textbf{CMOS} &4.13 & 2.98 & \textbf{3.81} & 3.46 & \underline{3.54} \\
\bottomrule
\end{tabular}
\vspace{-3mm}
\end{table}

\subsection{Analysis on Repeated Sampling}
\begin{figure}[t]
    \centering
    \includegraphics[width=0.46\textwidth]{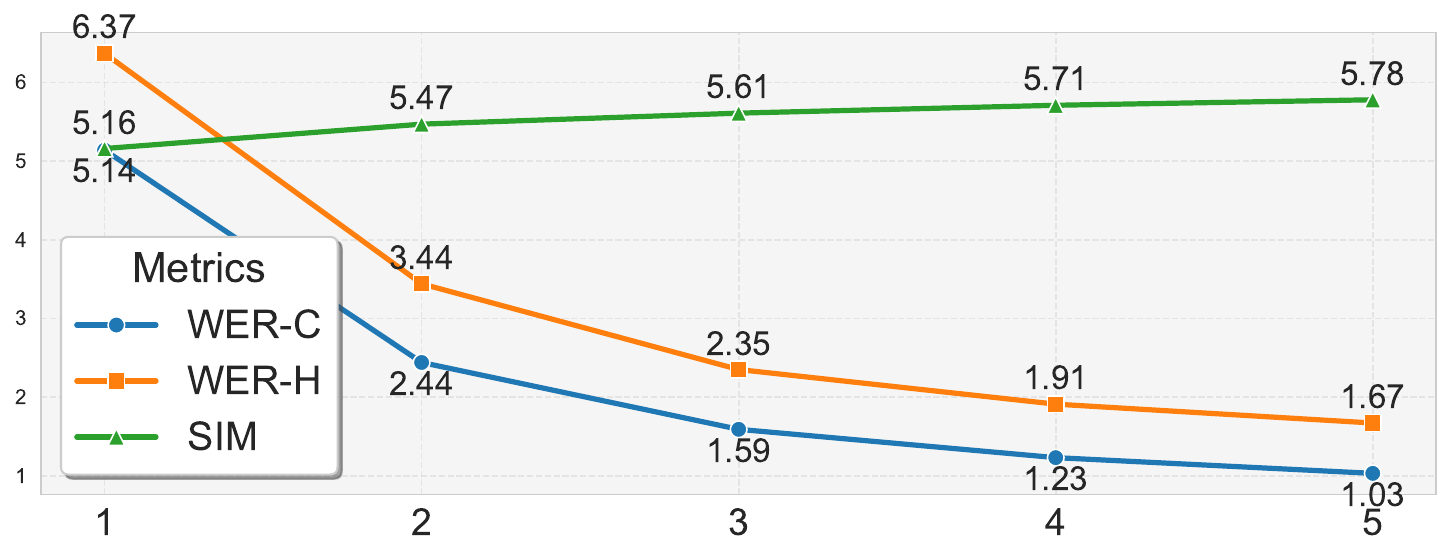}
    \vspace{-4mm}
    \caption{The repeated sampling ablation experiment, with the horizontal axis representing the number of repeated samplings. The SIM score is scaled by a factor of 10.}
    \label{fig:samplerepeat}
    \vspace{-7mm}
\end{figure}

To investigate how the Transducer model's output influences the SMLLE model and explore its full potential, the optimal results achieved at varying sampling frequencies during the five-sampling experiment, are illustrated in Figure \ref{fig:samplerepeat}.
It can be observed that as the number of samplings increases, both WER-C and WER-H steadily decrease, while SIM steadily increases, with the most significant improvements obtained after two samplings. The results clearly show that improving Transducer performance is key to optimizing SMLLE, thereby highlighting a clear pathway for future improvements.

\subsection{Analysis on DBM}
DBM enables the model to access future text information by removing \(\langle bos \rangle\) introducing as minimal delay as possible. As shown in Table \ref{dbm}, we analyze the effect of applying DBM to the prompt during the inference stage. It can be observed that the DBM application to prompts significantly degrades SMLLE's performance, as evidenced by increased WER scores. 
This may be due to the model not encountering the generation pattern of two sentences during training, which leads to difficulties in handling the two instances of DBM in both the prompt and the generated part. When DBM is not applied to the prompt, it provides a more stable initial state, which does not affect the generation process in the subsequent stages.

\begin{table}[ht]
\vspace{-3mm}
\centering
\caption{The impact of applying DBM to the prompt.}
\vspace{-3mm}
\label{dbm}
\begin{tabular}{cccc}
\toprule
Prompt DBM & WER-C & WER-H & SIM   \\ \midrule
\checkmark               & 10.26 & 13.28 & 0.487 \\
$\times$              & 5.14  & 6.37  & 0.516 \\ \bottomrule
\end{tabular}
\vspace{-5mm}
\end{table}

\section{Conclusion}
In this paper, we propose a novel zero-shot streaming TTS framework, SMLLE. It uses a Transducer model to convert text into semantic tokens in real time and reconstructs them into mel-spectrograms frame by frame using an AR model. We further design a DBM mechanism that allows SMLLE to access future text earlier, introducing as minimal delay as possible, thereby improving the stability of the model. Experimental results show that, in a streaming generation setting, SMLLE performs similarly to sentence-level TTS systems. In the same setting, SMLLE outperforms existing streaming TTS methods.



\bibliographystyle{IEEEtran}
\bibliography{mybib}

\end{document}